\documentclass[11pt]{article}

\usepackage{amsmath, amsfonts, amssymb, bm, bbm,color, amsthm, enumerate}

\usepackage{amsmath, amsfonts, amssymb, bm, bbm,color, amsthm, fullpage}
\usepackage[ruled, linesnumbered]{algorithm2e}

\usepackage{multirow}
\usepackage[authoryear, round]{natbib}
\usepackage{hyperref}

\usepackage{tikz}

\newcommand{\ep}{\mathbb{E}}

\usepackage{setspace}

\newcommand{\wass}{\mathrm{Wass}}

\usepackage{helvet}
 
\usepackage[T1]{fontenc}

\begin{document}

	\title{Consistent polynomial-time unseeded graph matching for Lipschitz graphons}
	\author{Yuan Zhang\\\\Department of Statistics,  The Ohio State University\\229 Cockins Hall, 1958 Neil Avenue, Columbus, Ohio, USA 43210\\\\yzhanghf@stat.osu.edu}
	
	\maketitle
	
	\begin{abstract}
		We propose a consistent polynomial-time method for the unseeded node matching problem for networks with smooth underlying structures.  Despite widely conjectured by the research community that the structured graph matching problem to be significantly easier than its worst case counterpart, well-known to be NP-hard, the statistical version of the problem has stood a challenge that resisted any solution both provable and polynomial-time.  The closest existing work requires quasi-polynomial time.  Our method is based on the latest advances in graphon estimation techniques and analysis on the concentration of empirical Wasserstein distances.  Its core is a simple yet unconventional sampling-and-matching scheme that reduces the problem from unseeded to seeded.  Our method allows flexible efficiencies, is convenient to analyze and potentially can be extended to more general settings.  Our work enables a rich variety of subsequent estimations and inferences.
	\end{abstract}
	
	\section{Introduction}
	In this paper, we study the graph matching problem formulated by:
	\begin{equation}
		\min_{P\in{\cal P}^{n\times n}}\|PW^{(1)}P^T - W^{(2)}\|_F
		\label{problem::graph_matching_oracle}
	\end{equation}
	where both $W^{(1)}$ and $W^{(2)}$ are $n\times n$ matrices and ${\cal P}^{n\times n}$ is the collection of all order-$n$ permutation matrices, and $\|\cdot\|_F$ is the Frobenius norm.  The prototype formulation \eqref{problem::graph_matching_oracle} may refer to either of the two main versions of this problem:  exact or inexact graph matching.  The exact graph matching problem suggests that we observe $W^{(1)}$ and $W^{(2)}$ directly, and suppose there is a true $P=P^*$ that perfectly matches them up by sending the Frobenius norm to 0.  This problem is well-known to be NP and current studies focus on upper bounding the worst case computational complexity.  The best known results belongs to \citet{babai2016graph}.  The inexact version of this problem only allows us to access noise-contaminated observations of $W^{(1)}$ and $W^{(2)}$ denoted by their adjacency matrices $A^{(1)}$ and $A^{(2)}$.  The inexact version usually comes with structural assumptions.  Despite this makes it a possibly easier problem than the exact matching problem without additional assumptions, it has stood unsolved by any consistent polynomial method.  In this paper, we focus on the inexact graph matching problem under mild structural assumptions.

	Most existing work on graph matching from accompanied by theoretical analysis from statistical perspectives assumed the true model to be stochastic block model \citep{holland1983stochastic}.  In this model, nodes are partitioned into $K$ non-overlapping communities, and each edge probability only depends on the memberships of the terminal nodes.  That is
	\begin{equation}
		W_{ij} = \ep[A_{ij}] = B_{c_i c_j}
		\label{def::SBM}
	\end{equation}
	where $W$ is the probability matrix and $B\in[0,1]^{K\times K}$ encodes community-level edge probabilities and $c=\{c_1,\ldots,c_n\}\in\{1,\ldots,K\}^n$ is the vector of node memberships.  Under a block model, the performance of an estimated node correspondence can be naturally measured by the number or proportion of nodes mapped to the same community in the other network.  Moreover, when the number of blocks is fixed or grows very slowly with the network size, with mild assumptions on the strength of the block structure, such as the network not being too sparse and the separation between within- and between-community edge probabilities being decently large, we can usually efficiently achieve consistent estimation of the node memberships by efficient methods such as spectral clustering \citep{von2007tutorial}.  Then by an exhaustive enumeration of all possible matchings between the communities of the two networks, we obtain an increasingly accurate estimated graph matching.  However, by far, theoretical guarantees for such approaches depend on assuming the true model is either exactly block model or its close variants such as the degree-corrected block model.  There is not yet any theory on the goodness of the output of a polynomial-time algorithm when the block model serves as a universal approximation to more general models as those described in \citet{bickel2009nonparametric, olhede2014network, gao2015rate, klopp2017oracle}, and \citet{choi2017co}.  It is straightforward to show that by optimal block model approximations to the adjacency matrices with properly selected block numbers followed by an optimal matching of the blocks that best aligns the two estimated block model approximations would achieve the minimax mean-squared error rate, but computing the optimal block model approximations per se require exponential computation time, and the subsequent step of aligning the blocks is again a difficulty problem.

	Beyond stochastic block models, many methods attempted to solve the noisy version of the problem:
	\begin{equation}
		\min_{P\in{\cal P}^{n\times n}}\|PA^{(1)}P^T - A^{(2)}\|_F
		\label{problem::graph_matching_sample}
	\end{equation}
	The obstacle is clearly the optimization over ${\cal P}^{n\times n}$ at a cardinality of $n!$, which is roughly comparable to $n^n$ by Stirling's formula.  A main line of efforts concentrated on relaxing $P$ to larger continuous spaces containing ${\cal P}^{n\times n}$ \citep{fishkind2012seeded, vogelstein2015fast, takapoui2016linear}, but it is commonly perceived challengingly difficult to analyze these relaxations and understanding their behaviors.  Moreover, negative results exist \citep{lyzinski2016graph}, suggesting that some ways of relaxation will almost never lead to desirable estimations.  Despite significant advances in recent years, the theoretical properties of solutions provided by relaxations remain largely unknown.  Recently, \citet{zhang2018unseeded} proposed a method suitable for low-rank graphon models by matching the estimated spectral embedding from the leading eigens of the adjacency matrices.  For more details about the embedding, see \citet{sussman2014consistent, levin2017central} or \citet{lei2018network}.  However, currently theory has only been established for low-rank populations.  Indeed, to consistently estimate the matching under a full-rank graphon, one may gradually increase the working rank of the estimation method in \citet{zhang2018unseeded} to achieve consistency, but the effectiveness of the inclusion procedure of increasingly many leading eigenvalues and eigenvectors depends on the eigen-structure of the underlying graphon, and despite the suggestion of close connections between graphon smoothness and its low-rankness \citep{udell2017nice}, currently no adaptive method is known.  In short, currently, no existing polynomial-time method could provide theoretically guaranteed consistency for full-rank graphons.
	
	We shall also briefly review a related topic that also contributed to inspiring this paper: the seeded graph matching problem.  Seeds refer to the set of known true correspondences between a potentially very small portion of nodes in the two networks.  They can be understood as anchor points in shape analysis  such as \citet{strait2017landmark}.  Knowing even just a few seeds can substantially lower the difficulty of the problem, reducing its complexity from NP to not only P, but usually an efficient computation time such as that demonstrated in \citet{lyzinski2015spectral} and \citet{mossel2018seeded}.  The seeded graph matching problem has been extensively studied and many empirical and theoretical results are known \citep{fishkind2012seeded, lyzinski2014seeded, lyzinski2016consistency}, especially under stochastic block models.
	
	An arguably common perception, as suggested in \citet{zhang2018unseeded} and possibly implicitly in \citet{fang2018tractable}, too, is that learning even one seed in the unseeded setting may be hard.  But in this paper, we will show that learning a set of ``soft'' seed nodes, where ``soft'' has similar meanings as that in \citep{fang2018tractable}, turns out to be easy.  Then under smoothness assumption, the ``soft'' seeds we select would gradually become ``hard'' as the bias asymptotically diminishes, and even randomly sampled sets of nodes can serve as quite good seed sets, as long as their numbers slowly grows to infinity.  This intuition is largely inspired by recent advances in concentration of empirical Wasserstein distances \citep{fournier2015rate, lei2018convergence}.  Essentially, our method reduces the unseeded graph matching problem from the scale of $n\times n$ to a much lower resolution at the scale of $(\log n/\log\log n)\times (\log n/\log\log n)$, where a dummy exhaustive enumeration of all possibilities becomes feasible.  Then as the resolution slowly enhances with the sample size, consistency is gradually achieved.
	
	\section{Set up}
	\label{section::set_up}
	
	In this paper, we focus on studying node-exchangeable networks.  Node exchangeability means an arbitrary permutation on the node order would not change the marginal distribution of the adjacency matrix $A$, see, for example, \citet{crane2018probabilistic}.  Under the exchangeability assumption, the network edge probability matrix $W$ can be re-expressed by the Aldous-Hoover representation \citep{aldous1981representations, hoover1979relations}:
	$$
	W_{ij} = \ep[A_{ij}] =  f(u_i, u_j)
	$$
	where $f$ is a latent function called the ``graphon function'' or simply ``graphon'', and $u_1,\ldots,u_n$ are latent node positions independently sampled from Uniform$(0,1)$.  Throughout this paper, we assume that the underlying graphon function $f$ is Lipschitz:
	\begin{equation}
	|f(x,y)-f(x',y')|\leq L(|x-x'|+|y-y'|)
	\label{assumption::Lipschitz}
	\end{equation}
	for a universal constant $L>0$.  For simplicity of narration, we mainly study symmetric and binary networks, that is, assume $f(x,y)=f(y,x)$ and $A_{ij}\sim$~Bernoulli$(W_{ij})$.  These two constraints are unessential, see the discussions in Section \ref{section::discussions}.
	
	Now we can formally set up the graph matching problem.  In this paper, we consider the matching problem between two ``matchable'' networks in the sense that they are generated from the same graphon function.  The latent positions and the induced probability matrices are given by
	\begin{align}
		\xi_1,\ldots,\xi_n&\stackrel{\textrm{iid}}{\sim} \textrm{Uniform}(0,1)\notag\\
		\eta_1,\ldots,\eta_n&\stackrel{\textrm{iid}}{\sim} \textrm{Uniform}(0,1)\notag\\
		W^{(1)}_{ij}&=f(\xi_1,\xi_j)\notag\\
		W^{(2)}_{ij}&=f(\eta_1,\eta_2)\notag
	\end{align}
	Here, the two sets of latent node positions $\{ \xi \}$ and $\{ \eta \}$ may be independent or dependent -- notice that this includes perfect dependence, in which case the two probability matrices would be identical at most up to a latent node permutation.  This dependency structure with more general correlation coefficients has been studied for Erd\"os-Renyi model by \citet{lyzinski2014seeded} and \citet{mossel2018seeded}.  Given the probability matrices $W^{(\ell)}$, the adjacency matrices are generated by
	\begin{equation}
		A_{ij}^{(\ell)}|W^{(\ell)}\sim \textrm{Bernoulli}\left(W^{(\ell)}_{ij}\right), \quad\ell=1,2
	\end{equation}
	and now we have fully described the data generation procedure.
	
	Next, it is important to properly define a goodness measure for performance evaluation.  Most existing graph matching paper assumed that there is a ``true'' node matching to be discovered, and their performance evaluations are naturally based on some discrepancy measures between the true and estimated node correspondences.  However, under our setting, the ``true'' mapping night not be a good choice for two reasons.  First, if the sets of latent node positions of the two networks are not perfectly dependent but instead partially correlated or even independent, then the  ``true'' node mapping does not exist; second, even if the two latent position sets are identical, due to the well-known model identifiability issues in graphon estimation \citep{airoldi2013stochastic}, there may exist other node correspondences that lead to identical data-generating distribution.  For example, consider a special stochastic block model called planted model, formulated as follows:
	$$
	W^{(1)} = W^{(2)} = ZBZ^T
	$$
	where
	$$
	B=\begin{pmatrix}
	p&q\\q&p
	\end{pmatrix}
	$$
	and $Z\in\{0,1\}^{n\times 2}$ that each row of $Z$ is either $(0,1)$ or $(1,0)$.  Then despite the underlying true node map is the identity map, any alternative map that sends all nodes in block 1 in network 1 to one block in network 2 and all nodes in block 2 to the other block in network 2 would induce identically distributed observable data.  For these reasons, similar to \citet{zhang2018unseeded}, we use the discrepancy between the two probability matrices, with nodes permuted by the estimated correspondence, as the goodness measure.  The loss function is formulated as follows:
	\begin{equation}
		\textrm{Loss}(\widehat{P}) = \frac1n\left\|\widehat{P}W^{(1)}\left( \widehat{P} \right)^T - W^{(2)}\right\|_F
		\label{def::goodness_measure}
	\end{equation}
	
	\section{Our method}
	
	We start by outlining the intuitions of our method.  In the unseeded setting we do not have any clue of seed nodes.  We crack this obstacle by manually cranking out a few seed nodes by simply randomly sampling them from all nodes.  From the recent advances in empirical Wasserstein distance theory \citep{fournier2015rate, lei2018convergence}, we know that as more are sampled as artificial seeds, the seed node set would become increasingly representative in the sense by asymptotically covers all the connection behavior patterns in the network they are sampled from.  Essentially, sampling from all nodes can be viewed as sampling latent positions from Uniform$(0,1)$ by recalling the Aldous-Hoover representation, and the set of their latent positions become ``denser'' in the sample space $(0,1)$.  Therefore in the limit, for any node positioned at $u\in(0,1)$, with high probability, there exists a node in the artificial seed set with position $u'$ close to $u$.  Their graphon slices $f(\cdot, u)$ and $f(\cdot, u')$ will also be close due to the graphon smoothness assumption.  Moreover, the latent node positions of the two artificial seed sets would approach each other growingly well in $\ell_2$ Wasserstein distance.  We can first reduce the problem of matching all nodes in the two networks to matching their representatives, and second, we may contain the computational complexity of the reduced optimization problem by letting the sizes of the two artificial seed node sets grow sufficiently slow.
	
	At a high level, our method employs a block model approximation to graphon estimation, but it has important differences from those in existing literature.  Currently, there are two main directions in the study of block models.  One is community detection, in which researchers have been pushing the theories in aspects such as detection limit, rate-optimality, computational efficiency and so on, to their limits, see \citet{zhao2017survey} for a comprehensive survey; however, most community detection theory critically rely on the exact piece-wise constant model assumption, and many of them focused on special cases such as the planted model and do not easily extend to more general ``block-like'' models such as that in \citet{guedon2016community}.  The other research direction of is block model approximations to general graphons with global or piece-wise smoothness assumptions.  Unfortunately, to the author's best knowledge, existing methods along this line lacks either computational feasibility or theoretical guarantees.   There is apparently a blank in available tools that both feasibly and provably approximate the true model with block model, and our work provides the first candidate in the attempt to bridge this gap.  When using block models to approximate general graphons, one may no longer assume separability or signal-to-noise conditions as that in community detection, to which community detection theories largely held on tightly to; on the other hand, these conditions for community detection are usually unimportant for graphon estimation.  For an illustrative example of this difference, see the discussions in \citet{gao2015rate}.
	
	Yet another line of matrix denoising methods are based on singular value decomposition without using blocks \citep{chatterjee2015matrix, klopp2017optimal}, but it is unclear how they could help to substantially reduce the scale for the graph matching problem.  One substantive difference between SVD-based methods and block model approximation is that the former does not provide a ``re-organization'' of nodes into some estimated clusters and the latter does.  This is very slight difference in graphon estimation for analyzing a single network, and an estimated node ordering only contributes to interpretability and does not help the performance evaluations that usually focused on MSE.  But in the graphon matching context, grouping nodes together is a viable way to reduce the problem scale, giving the block model approximation approach arguable preferences over SVD-based methods.
	
	Now let us formally introduce our method.  The input data are $A^{(1)}$ and $A^{(2)}$, generated from their underlying edge probability matrices $W^{(1)}$ and $W^{(2)}$, respectively.  Recall that there might or might not exist a ground-truth permutation matrix $P^*\in{\cal P}^{n\times n}$ that perfectly aligns $W^{(1)}$ to $W^{(2)}$, and even when a well-defined $P^*$ does exist, our goal is not to recover it because its possible lack of uniqueness but instead suppressing the loss in \eqref{def::goodness_measure} to 0 asymptotically with a polynomial-time estimator $\widehat{P}$.  Our designed algorithm is as follows:
	
	\begin{algorithm}[H]
		\caption{Unseeded graph matching}
		\label{algorithm::main}
		\KwIn{Observed shuffled adjacency matrices $A^{(1)}, A^{(2)}\in\mathbb{R}^{n\times n}$ and artificial seed set size $d$.  Without loss of generality assume $d|n$ (otherwise randomly eliminate $n\%d$ nodes).}
		\KwOut{Estimated permutation matrix $\widehat{P}\in{\cal P}^{n\times n}$}
		Denoise the columns of $A^{(\ell)}, \ell=1,2$ by Neighborhood Smoothing \citep{zhang2017estimating} and obtain $\widehat{W}^{(\ell)}, \ell=1,2$\;
		Randomly choose $d$ columns in $\widehat{W}^{(1)}$ and $d$ columns in $\widehat{W}^{(2)}$, independently.  Denote their indices of the selected columns as ${\cal I}:=\{i_1,\ldots,i_d\}$ and ${\cal J}:=\{j_1,\ldots,j_d\}$\;
		Set $$
		R = \begin{pmatrix}
		\mathbbm{1}_{n/d}^T & 0 & 0 \cdots 0& 0\\
		0 & \mathbbm{1}_{n/d}^T & 0 \cdots 0& 0\\
		\cdots&\cdots&\cdots & \cdots\\
		0 & 0 & 0\cdots 0 & \mathbbm{1}_{n/d}^T
		\end{pmatrix}
		$$ and define $\breve{W}^{(1)} = \widehat{W}^{(1)}_{\cdot,{\cal I}} R$ and $\breve{W}^{(2)} = \widehat{W}^{(2)}_{\cdot,{\cal J}} R$.  Then solve
		\begin{align*}
			\widehat{P}^{(1)}&:=\arg\min_{P\in{\cal P}^{n\times n}}\left\| \widehat{W}^{(1)}P^T - \breve{W}^{(1)} \right\|_F\\
			\widehat{P}^{(2)}&:=\arg\min_{P\in{\cal P}^{n\times n}}\left\| \widehat{W}^{(2)}P^T - \breve{W}^{(2)} \right\|_F
		\end{align*}
		by running a Hungarian algorithm on the columns of the corresponding matrices\;
		Solve the optimization problem
		\begin{equation}
			\widehat{Q}:=\arg\min_{Q=I_{n/d}\otimes \widetilde{Q}: \widetilde{Q}\in{\cal P}^{d\times d}}\left\| Q\widehat{P}^{(1)}\widehat{W}^{(1)}\left( \widehat{P}^{(1)} \right)^TQ^T - \widehat{P}^{(2)}\widehat{W}^{(2)}\left( \widehat{P}^{(2)} \right)^T \right\|_F
		\end{equation}
		where $\otimes$ is the Kronecker product, by exhaustively enumerating all possible $\widetilde{Q}$'s
		\label{step::step_5}\;
		Return $\widehat{P}:=\left(\widehat{P}^{(2)}\right)^T\widehat{Q}\widehat{P}^{(1)}$.
	\end{algorithm}
	
	Now we calculate the time complexity.  Clearly all steps are at most $O(n^3)$ except for Step \ref{step::step_5}, and Step \ref{step::step_5} costs $O(d! \cdot n^2)$.  Taking
	$$
	d= C \frac{\log n}{\log\log n}
	$$
	we have that
	$$
	d!\asymp d^{d+1/2}/e^d\preceq d^d = e^{d\log d} = e^{C \cdot \frac{\log n}{\log\log n}(\log\log n - \log\log\log n)}\asymp n^{C}
	$$
	for some universal constant $C>0$.  This makes the proposed method polynomial-time.

	\section{Proof of consistency}
	
	This section is dedicated to the proof that under the smoothness assumption, $\widehat{P}$ estimated by Algorithm \ref{algorithm::main} is consistent in the sense that
	$$
	\frac1n\left\| \widehat{P}W^{(1)}\left( \widehat{P} \right)^T - W^{(2)} \right\|_F \stackrel{P}{\longrightarrow}0
	$$
	We start the proof by recalling the theory of \citet{zhang2017estimating}.  With high probability, we have
	\begin{equation}
		\max_{k=1,\ldots,n; \ell=1,2} \frac1{\sqrt{n}}\left\| \widehat{W}_{\cdot,k}^{(\ell)} - W_{\cdot,k}^{(\ell)} \right\|_2 \leq C\sqrt{\frac{\log n}{n}}
		\label{eqn::NS}
	\end{equation}
	We now analyze the partition of nodes in network 1 into $d$ clusters concentrated around the $d$ randomly sampled seed nodes.  Similar analysis holds for network 2 and will be omitted.  Recall that the latent node positions for network 1 are $\xi_1,\ldots,\xi_n$ and those for network 2 are $\eta_1,\ldots,\eta_n$, and the latent positions of the artificial seed nodes are $\xi_{i_1},\ldots,\xi_{i_d}$ and $\eta_{j_1},\ldots,\eta_{j_d}$, respectively.  Define the latent map that finds the $\ell_2$ Wasserstein distances between $(\xi_1,\ldots,\xi_n)$ and $(\xi_{i_1},\ldots,\xi_{i_d}) \cdot R$ and that between $(\eta_1,\ldots,\eta_n)$ and $(\eta_{j_1},\ldots,\eta_{j_d})\cdot R$ by
	\begin{align*}
		P^{*(1)}&:=\arg\min_{P\in{\cal P}} \left\| (\xi_1,\ldots,\xi_n) P^T - (\xi_{i_1},\ldots,\xi_{i_d}) \cdot R \right\|_2\\
		P^{*(2)}&:=\arg\min_{P\in{\cal P}} \left\| (\eta_1,\ldots,\eta_n) P^T - (\eta_{j_1},\ldots,\eta_{j_d}) \cdot R \right\|_2
	\end{align*}
	Notice that both $P^{*(\ell)},\ell=1,2$ are not estimable due to the contiguity on $\xi$ and $\eta$'s.  By \citet{levina2001earth,fournier2015rate} and \citet{lei2018convergence}, with high probability, we have
	\begin{align}
		&\frac1{\sqrt{n}}\left\| (\xi_1,\ldots,\xi_n) \left( P^{*(1)} \right)^T - (\xi_{i_1},\ldots,\xi_{i_d}) \cdot R \right\|_2\notag\\
		=& \wass\left( {\cal F}(\xi_1,\ldots,\xi_n), {\cal F}((\xi_{i_1},\ldots,\xi_{i_d}) \cdot R) \right)\notag\\
		=& \wass\left( {\cal F}(\xi_1,\ldots,\xi_n), {\cal F}((\xi_{i_1},\ldots,\xi_{i_d})) \right)\leq C d^{-1/2}
		\label{eqn::Lipschitz_1}
	\end{align}
	where $\wass$ refers to the $\ell_2$ Wasserstein distance and ${\cal F}(x_1,\ldots,x_n)$ denotes the empirical distribution induced by the point set $\{ x_1,\ldots,x_n \}$, and notice that ${\cal F}((\xi_{i_1},\ldots,\xi_{i_d})R) = {\cal F}(\xi_{i_1},\ldots,\xi_{i_d})$.  An exactly similar inequality can be proved for network 2.  Then defining $\widetilde{W}^{(\ell)} := W^{(\ell)}R$, by \eqref{eqn::Lipschitz_1} and the Lipschitz assumption \eqref{assumption::Lipschitz}, with high probability, we have
	\begin{equation}
		\left\| W^{(\ell)} \left( P^{*(\ell)} \right)^T - \widetilde{W}^{(\ell)} \right\|_F \leq  L \left\| (\xi_1,\ldots,\xi_n) \left( P^{*(1)} \right)^T - (\xi_{i_1},\ldots,\xi_{i_d}) \cdot R \right\|_2\leq   C n d^{-1/2}
		\label{eqn::pop_bias}
	\end{equation}
	From this point on, we abuse the notation of universal constants $C$ in the sense that the same symbol $C$ may represent different universal constants from line to line.  By \eqref{eqn::pop_bias}, with high probability, we have:
	\begin{align}
		&\left\| W^{(\ell)}\left( \widehat{P}^{(\ell)} \right)^T - \widetilde{W}^{(\ell)} \right\|_F  \leq  \left\| \widehat{W}^{(\ell)}\left( \widehat{P}^{(\ell)} \right)^T - \breve{W}^{(\ell)} \right\|_F + C\sqrt{n\log n}\notag\\
		\leq&  \left\| \widehat{W}^{(\ell)}\left( P^{*(\ell)} \right)^T - \breve{W}^{(\ell)} \right\|_F + C\sqrt{n\log n}  \leq \left\| W^{(\ell)}\left( P^{*(\ell)} \right)^T - \widetilde{W}^{(\ell)} \right\|_F  + 2C\sqrt{n\log n}\notag\\
		\leq& Cnd^{-1/2} + C\sqrt{n\log n}\asymp Cnd^{-1/2}
		\label{eqn::control_1}
	\end{align}
	From \eqref{eqn::pop_bias} and \eqref{eqn::control_1}, we have
	\begin{align}
		&\left\| W^{(\ell)}\left( \widehat{P}^{(\ell)} \right)^T - W^{(\ell)}\left( P^{*(\ell)} \right)^T \right\|_F \notag\\
		\leq & \left\| W^{(\ell)}\left( \widehat{P}^{(\ell)} \right)^T - \widetilde{W}^{(\ell)} \right\|_F + \left\| W^{(\ell)}\left( P^{*(\ell)} \right)^T - \widetilde{W}^{(\ell)} \right\|_F \leq  C nd^{-1/2}
		\label{eqn::control_2}
	\end{align}
	Then using \eqref{eqn::control_2}, we have
	\begin{align}
		&\left\| \widehat{P}^{(\ell)} W^{(\ell)}\left( \widehat{P}^{(\ell)} \right)^T - P^{*(\ell)}W^{(\ell)}\left( P^{*(\ell)} \right)^T \right\|_F\notag\\
		\leq & \left\| \widehat{P}^{(\ell)}\left\{ W^{(\ell)}\left(\widehat{P}^{(\ell)}\right)^T - W^{(\ell)}\left( P^{*(\ell)} \right)^T \right\} \right\|_F + \left\| \left\{ \widehat{P}^{(\ell)}W^{(\ell)} - P^{*(\ell)}W^{(\ell)} \right\}  \left( P^{*(\ell)} \right)^T \right\|_F\notag\\
		=& 2\left\| W^{(\ell)}\left( \widehat{P}^{(\ell)} \right)^T - W^{(\ell)}\left( P^{*(\ell)} \right)^T \right\|_F  \leq Cnd^{-1/2}
		\label{eqn::control_2.5}
	\end{align}
	By the graphon Lipschitz assumption, with high probability, we have
	\begin{align}
		&\left\| P^{*(1)}W^{(1)}\left( P^{*(1)} \right)^T - \left(\mathbbm{1}_{n/d}\mathbbm{1}_{n/d}^T\right)\otimes W_{{\cal I}, {\cal I}}^{(1)}  \right\|_F \notag\\
		\leq & C\left[\left\| P^{*(1)} \begin{pmatrix}
		\xi_1\\\vdots\\\xi_n
		\end{pmatrix} \mathbbm{1}_n^T \left( P^{*(1)} \right)^T  -  R^T\begin{pmatrix}
		\xi_{i_1}\\\vdots\\\xi_{i_d}
		\end{pmatrix} \mathbbm{1}_n^T  \right\|_F^2  +  \left\| P^{*(1)}\mathbbm{1}_n (\xi_1,\ldots,\xi_n) \left( P^{*(1)}\right)^T - \mathbbm{1}_n (\xi_{i_1},\ldots\xi_{i_d})R \right\|_F^2\right]^{1/2}\notag\\
		\leq & C\left\{  2n\left\| (\xi_1,\ldots,\xi_n)\left( P^{(1)} \right)^T - (\xi_{i_1},\ldots,\xi_{i_d})R \right\|_2^2  \right\}^{1/2} \leq Cnd^{-1/2}
		\label{eqn::control_3}
	\end{align}
	Similarly, one can show that with high probability,
	\begin{equation}
		\left\| P^{*(2)}W^{(2)}\left( P^{*(2)} \right)^T - \left(\mathbbm{1}_{n/d}\mathbbm{1}_{n/d}^T\right)\otimes W_{{\cal J}, {\cal J}}^{(2)}  \right\|_F\leq Cnd^{-1/2}
		\label{eqn::control_3.5}
	\end{equation}
	
	Define the oracle permutation matrix $\widetilde{Q}^*\in{\cal P}^{d\times d}$ to be the latent mapping that finds the $\ell_2$ Wasserstein distance between $(\xi_{i_1},\ldots,\xi_{i_d})$ and $(\eta_{j_1},\ldots,\eta_{j_d})$.  That is,
	\begin{equation}
		\widetilde{Q}^*:=\arg\min_{Q\in{\cal P}^{d\times d}} \left\| (\xi_{i_1},\ldots,\xi_{i_d}) Q^T - (\eta_{j_1},\ldots,\eta_{j_d}) \right\|_2
		\label{eqn::Lipschitz_2}
	\end{equation}
	Then similar to the derivations in \eqref{eqn::control_3}, by the Lipschitz assumption, \eqref{eqn::Lipschitz_2} implies
	\begin{equation}
		\frac1d\left\| \widetilde{Q}^* W^{(1)}_{{\cal I},{\cal I}}\left( \widetilde{Q}^* \right)^T - W^{(2)}_{{\cal J},{\cal J}} \right\|_F \leq Cd^{-1/2}
		\label{eqn::control_4}
	\end{equation}
	Defining $Q^*:=I_{n/d}\otimes \widetilde{Q}^*$, \eqref{eqn::control_4} is equivalent to
	\begin{align}
		\left\| Q^* \left\{\left( \mathbbm{1}_{n/d}\mathbbm{1}_{n/d}^T \right)\otimes W^{(1)}_{{\cal I},{\cal I}}\right\}\left( Q^* \right)^T - \left( \mathbbm{1}_{n/d}\mathbbm{1}_{n/d}^T \right)\otimes W^{(2)}_{{\cal J},{\cal J}} \right\|_F \leq Cnd^{-1/2}
		\label{eqn::pop_bias_2}
	\end{align}
	since the elements in every $n/k$ by $n/k$ block in $Q^* \left\{\left( \mathbbm{1}_{n/d}\mathbbm{1}_{n/d}^T \right)\otimes W^{(1)}_{{\cal I},{\cal I}}\right\}\left( Q^* \right)^T$ and $\left( \mathbbm{1}_{n/d}\mathbbm{1}_{n/d}^T \right)\otimes W^{(2)}_{{\cal J},{\cal J}}$ are copies of the corresponding elements in $\widetilde{Q}^* W^{(1)}_{{\cal I},{\cal I}}\left( \widetilde{Q}^* \right)^T$ and $W^{(2)}_{{\cal J},{\cal J}}$, respectively.
	
	We are now ready to complete the proof.  We have
	\begin{align}
		&\left\|  \widehat{Q}\widehat{P}^{(1)} W^{(1)} \left( \widehat{P}^{(1)} \right)^T\widehat{Q}^T - \widehat{P}^{(2)}W^{(2)}\left( \widehat{P}^{(2)} \right)^T  \right\|_F\notag\\
		\leq& \left\|  \widehat{Q}\widehat{P}^{(1)} \widehat{W}^{(1)} \left( \widehat{P}^{(1)} \right)^T\widehat{Q}^T - \widehat{P}^{(2)}\widehat{W}^{(2)}\left( \widehat{P}^{(2)} \right)^T  \right\|_F + C\sqrt{n\log n} \notag\\
		\leq& \left\|  Q^*\widehat{P}^{(1)} \widehat{W}^{(1)} \left( \widehat{P}^{(1)} \right)^T\left(Q^*\right)^T - \widehat{P}^{(2)}\widehat{W}^{(2)}\left( \widehat{P}^{(2)} \right)^T  \right\|_F + C\sqrt{n\log n} \notag\\
		\leq& \left\|  Q^*\widehat{P}^{(1)} W^{(1)} \left( \widehat{P}^{(1)} \right)^T\left(Q^*\right)^T - \widehat{P}^{(2)}W^{(2)}\left( \widehat{P}^{(2)} \right)^T  \right\|_F + C\sqrt{n\log n} \notag\\
		\leq& \left\|  Q^* P^{*(1)} W^{(1)}\left(P^{*(1)}\right)^T  \left(Q^*\right)^T  -  P^{*(2)}W^{(2)}\left( P^{*(2)}  \right)^T  \right\|_F + Cnd^{-1/2} + C\sqrt{n\log n} \notag\\
		\leq& \left\|  Q^* \left\{ \left( \mathbbm{1}_{n/d}\mathbbm{1}_{n/d}^T \right)\otimes W^{(1)}_{{\cal I},{\cal I}} \right\}  \left(Q^*\right)^T  -  \left( \mathbbm{1}_{n/d}\mathbbm{1}_{n/d}^T \right)\otimes W^{(2)}_{{\cal J},{\cal J}}  \right\|_F + Cnd^{-1/2} + C\sqrt{n\log n} \notag\\
		\leq & Cnd^{-1/2}
	\end{align}
	holds with high probability, where in Lines 2 and 4, we used \eqref{eqn::NS}, in Line 3, we used the definition of $\widehat{Q}$, in Line 5, we used \eqref{eqn::control_2.5}, in Line 6, we used \eqref{eqn::control_3} and \eqref{eqn::control_3.5}, and finally in Line 7, we used \eqref{eqn::pop_bias_2}.  The proof of consistency is then finalized by recalling that we set $d\asymp \log n/\log\log n$.

	\section{Discussions}
	\label{section::discussions}
	
	We first discuss variants and possible extensions of our method.  We believe that our method and its theoretical guarantee can be extended to piece-wise Lipschitz graphons, and this will include stochastic block models.  Also we conjecture the current results to hold for not only binary, but more generally sub-Gaussian adjacency matrix element generation schemes.  Our method presented in this paper is designed for symmetric networks, and in fact handling asymmetric networks are not harder.  We explain for two kinds of asymmetric networks.  The first type is bipartite networks, in which the sets of edge senders and edge receivers are two potentially unrelated sets.  In Aldous-Hoover representation, they are written as $W_{ij} = f(\xi_i, \tilde{\xi}_j)$ where $W\in\mathbb{R}^{m\times n}$.  To match bipartite networks, we simply modify Step \ref{step::step_5} as follows: after $\widehat{P}^{(\ell)}, \ell=1,2$ are estimated, one may simply run a Hungarian algorithm to match up the rows of $\widehat{W}^{(1)}\left( \widehat{P}^{(1)} \right)^T$ to the rows of $\widehat{W}^{(2)}\left( \widehat{P}^{(2)} \right)^T$, and the theoretical analysis turns out to be even easier than the symmetric case.  In the second type of asymmetric networks, participating nodes are the same set of individuals, but the edges are directed.  Depending on whether the assumption that nodes behaving similarly as senders also behave similarly as receivers is reasonable or not, we may use apply the methods for symmetric (if the assumption is true) or bipartite networks, and the corresponding theories should follow.
	
	On the technical side, it is of interest to seek improvements in the error rate while staying computationally feasible.  However, as mentioned earlier, if we do not demand computational feasibility, the minimax mean-squared error rate under our model assumptions is identical to that for matrix estimation.  To achieve the optimal error rate, simply use network histograms \citep{olhede2014network} with equal numbers of bins, with the common bin number chosen by the theory of \citet{klopp2017oracle}, to approximate the two network, respectively.  Then take the optimal permutation on all blocks of one of the block model approximations that minimizes its discrepancy in Frobenius norm with the other would induce a desired estimated matching.  Similar results also hold for piece-wise Lipschitz graphon by slightly varying the proof Proposition 1 in \citet{zhang2017estimating}.
	
	Polynomial-time consistent estimation for unseeded graphon estimation enables rich future inferences, such as network comparison \citep{asta2014geometric, tang2017nonparametric, lei2018network}, independence test and correlation estimation \citep{lyzinski2018information}, change point detection \citep{wang2017fast}, joint estimations of common structural parameters by combining information from several networks \citep{le2017estimating, liu2018global} and so on.  It is also of both theoretical and practical interest to analyze the asymptotic limiting distribution for  full-rank graphons, analogous to existing results under low-rank models including \citet{tang2016limit} and \citet{levin2017central}.
	
	Before concluding this paper, let us mention two technical details.  First, we have confined ourselves within networks of equal sizes.  This can be easily overcome by sampling without replacements the nodes of the larger network and match the induced network to the smaller network.  The remaining nodes in the larger network can then join the estimated node correspondence through the estimated clustering structure.  We can roughly think the randomly selected seed nodes in the larger network as ``cluster centers'' and estimate the clustering structure by running a Hungarian algorithm on columns denoised by neighborhood smoothing, similar to that in \eqref{eqn::Lipschitz_1}.  The sending every left-over node $i$ in the larger network to an arbitrary node in the smaller network that is clustered to the artificial seed node corresponding to the seed node in the large network that $i$ is clustered to.  The second topic is the incorporation of potentially available seed nodes.  One naive adjustment is to simply replace part of randomly sampled artificial seed nodes by the known ones, and set the known correspondence between the available seed nodes fixed in Step \ref{step::step_5}.  However, it is entirely possible that seed nodes can be used in a much more effective way that substantively improves the error rate, and this is certainly an interesting future research direction.

	\bibliographystyle{abbrvnat}
	\bibliography{Manuscript_5}

\end{document}